\documentclass[letterpaper, 10 pt, journal, twoside]{IEEEtran}
\usepackage[utf8]{inputenc}
\usepackage{amsmath,amssymb,amsfonts}
\usepackage{bm}
\usepackage{url}
\usepackage{graphicx,graphics}
\usepackage{multirow}
\usepackage{booktabs}
\usepackage{multirow}
\usepackage[ruled,linesnumbered]{algorithm2e}
\usepackage{textcomp}
\usepackage{tabularx}
\usepackage{relsize}
\usepackage[flushleft]{threeparttable}
\usepackage{cite}
\usepackage[table,dvipsnames]{xcolor}
\newcommand{\midsepremove}{\aboverulesep = 0.2mm \belowrulesep = 0.2mm}
    \midsepremove
\newcommand{\midsepdefault}{\aboverulesep = 0.605mm \belowrulesep = 0.984mm}
    \midsepdefault
\title{Simultaneous Tactile Exploration and Grasp Refinement for Unknown Objects}
\author{{Cristiana de Farias, Naresh Marturi, Rustam Stolkin, Yasemin Bekiroglu}
\thanks{Manuscript received: October 15, 2020; Revised: January 8, 2021; Accepted: February 7, 2021.}
\thanks{This letter was recommended for publication by Editor Hong Liu upon evaluation of the Associate Editor and Reviewers’ comments.}%
\thanks{This work was supported by the UK National Centre for Nuclear Robotics (NCNR), Chalmers AI Research Center (CHAIR) and Chalmers Gender Initiative for Excellence (Genie). Part funded by CHIST-ERA under Project EP/S032428/1 PeGRoGAM and in part supported by Faraday Institution sponsored Recycling of Lithium Ion Batteries (ReLiB) project (grant: FIRG005). \textit{(Corresponding Author: Cristiana de Farias)}}%
\thanks{C. de Farias, N. Marturi, and R. Stolkin are with the Extreme Robotics Laboratory, School of Metallurgy and Materials, University of Birmingham, Birmingham, U.K.  (email: CXM1029@student.bham.ac.uk; n.marturi@bham.ac.uk; r.stolkin@bham.ac.uk).}%
\thanks{Y. Bekiroglu is with Chalmers University of Technology, Department of Electrical Engineering, Automatic Control research group, Sweden and University College London, Department of Computer Science, Centre for Artificial Intelligence, U.K. (email:yaseminb@chalmers.se, y.bekiroglu@ucl.ac.uk).}%
\thanks{Digital Object Identifier (DOI): see top of this page.}
}
\begin{document} \sloppy
\bstctlcite{IEEEexample:BSTcontrol}
\maketitle
\begin{abstract}
This paper addresses the problem of simultaneously exploring 
an unknown object to model its shape, using tactile sensors on robotic fingers, while also improving finger placement to optimise grasp stability. In many situations, a robot will have only a partial camera view of the near side of an observed object, for which the far side remains occluded. We show how an initial grasp attempt, based on an initial guess of the overall object shape, yields tactile glances of the far side of the object which enable the shape estimate and consequently the successive grasps to be improved. We propose a grasp exploration approach using a probabilistic representation of shape, based on Gaussian Process Implicit Surfaces. This representation enables initial partial vision data to be augmented with additional data from successive tactile glances. This is combined with a probabilistic estimate of grasp quality to refine grasp configurations. When choosing the next set of finger placements, a bi-objective optimisation method is used to mutually maximise grasp quality and improve shape representation during successive grasp attempts. Experimental results show that the proposed approach yields stable grasp configurations more efficiently than a baseline method, while also yielding improved shape estimate of the grasped object.
\end{abstract}%

%
\section{Introduction}
\label{sec:intro}
\IEEEPARstart{H}{umans} often begin a reach-to-grasp motion with only a brief visual glimpse of the object being grasped. Once our hand makes contact with the object, we exploratively wrap our fingers around it, progressively modifying and refining a stable grasp, while also improving our mental model of the object's 3D structure, using extremely rich tactile information from the human hand and fingers. Studies show that humans combine visual and tactile information in optimal ways for perception of objects \cite{ernst}; and also when planning and executing grasp and manipulation actions, also triggering plan corrections based on predictions \cite{johansson,johansson2}. However, transferring those skills to robots, to achieve safe and robust grasping, remains a major challenge. A key technical difficulty is how to encode, reason about and overcome the uncertainties that are inherent in both robotic perception and also in the robot's physical interactions with objects and surfaces.

A variety of approaches to grasping and manipulation have been proposed, which tackle uncertainty about diverse object properties that are not fully observable \textit{a-priori}, such as shape, pose, friction, inertia and mass distribution \cite{gpg,berkeley2DGPIS,graspGPIS}. One of the most important object properties, in terms of its influence on the way we grasp, is shape  \cite{humangrasp, neuroscience}. A good representation of object shape should ideally allow for: i) encoding uncertainty about the shape, with uncertainty varying over different surface regions; ii) optimal fusion of different sources of information, including e.g. tactile and visual; iii) incremental improvement and modification of the shape model over time, with successive sensory inputs; iv) supporting probabilistic estimate of success in different grasp poses, and the generation of fluent reach-to-grasp motions.
\begin{figure}
\centering
    \includegraphics[width=0.98\columnwidth]{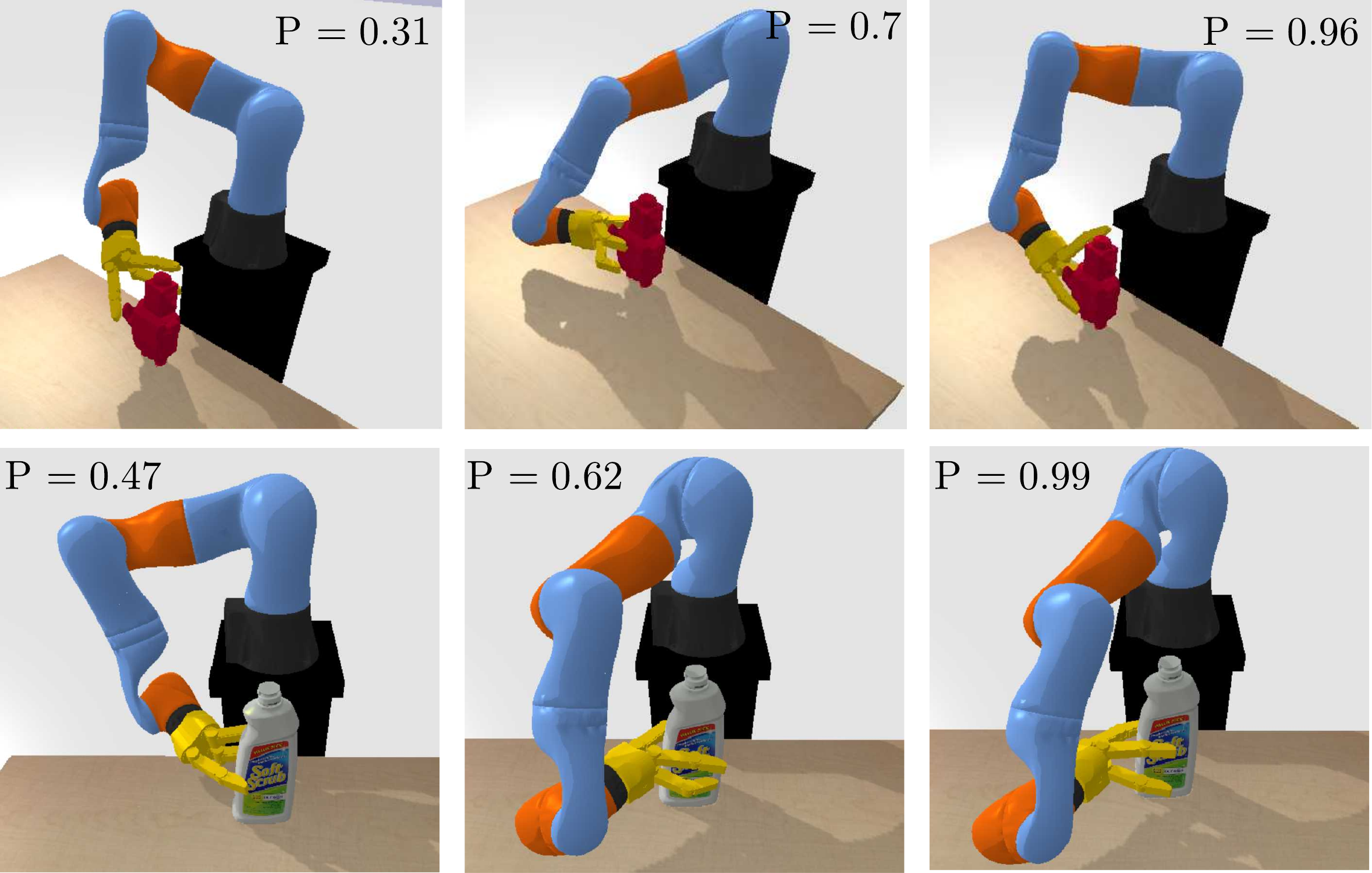}
    \caption{Tactile exploration and grasp refinement of two unknown objects. Scores denoted by ``P'' are the success probabilities at various iterations.}
  \label{fig:GPISBO_FullArm}
\end{figure}

A probabilistic shape representation, notably Gaussian process implicit surfaces (GPIS) \cite{microsoftPaper}, which can be built from sensory data \cite{bjorkman, gpis_journal_bekiroglu, stanimir_shape_grasp}, is a good candidate to address above requirements. They have been shown to yield sufficient surface reconstructions to identify or categorize objects \cite{bjorkman, gpis_journal_bekiroglu}. In this paper, we show how GPIS-based shape modelling can be used to generate stable grasp configurations, given partial point clouds of \textit{a-priori} unknown objects, in combination with an efficient tactile exploration strategy. %

Planning reach-to-grasp motions which, when executed, will reliably result in stable grasps actually being achieved, is challenging and faced with fundamental limitations. A good motion plan can be disrupted by numerous sources of small errors and uncertainties, which can accumulate catastrophically during open-loop execution. Nothing can be perfectly modelled, with diverse types of additive uncertainty including: kinematic, dynamic and stiffness parameters of high-dof arms and hands; shape and 6-dof pose of the object being grasped; calibration errors between the robot and vision system; noise and distortions in visual and other sensory reconstruction data; material properties affecting frictional and contact forces. Without sensory feedback during task execution, and reactive online plan corrections, original reach-to-grasp plan often fails.

To this end, we propose an approach to grasping which combines pre-execution planning with online sensing and re-planning in a principled way. Our method is initialised  using a partial point cloud of the object, captured from a single depth-camera view. This is used to plan an initial grasp configuration of the hand and a reach-to-grasp trajectory. We show how a probabilistic object shape representation constructed as a GPIS can be used in a Bayesian Optimisation (BO) approach to plan grasps with a high probability of success.
We propose a bi-objective optimisation framework, in which successive finger placements are planned to \textit{mutually improve} both grasp stability and shape representation. Our approach enables  tactile exploration for simultaneously refining grasps and perceived shape iteratively. We present a simulated robot with a dexterous hand as in Fig. \ref{fig:GPISBO_FullArm}. The proposed approach yields stable grasp configurations more efficiently than a baseline method, improving the shape estimate of the grasped object.
%
\section{Related Work}
\label{sec:related}
Autonomous robotic grasping of \textit{a-priori} unknown objects (\textit{i.e.}, methods that do not rely on object CAD models) has become a significant research topic, with many approaches proposed in recent years. However, a robust and generalisable grasp-planning solution remains elusive \cite{survey_unknown}. As an alternative to precise grasp planning computation, with complex fully actuated robotic hands, under-actuated gripper designs have also been proposed, which mechanically adapt to different object shapes \cite{bicchi}. Grasp planning algorithms for unknown objects can be categorized into global grasping approaches \cite{box-based} considering the whole object to find the best grasp, and local grasping approaches \cite{local,locomo} using partial data from the object, e.g. local contact moments \cite{locomo}, or a hierarchical representation using edge and texture information to generate grasps from only visible parts of objects that may cause failures \cite{local}. In general, global approaches are preferred for multi-finger grasping \cite{survey_unknown}.

Choice of object representation has a direct effect on the characteristics, efficiency and performance of a planner. 
Coarse approximations of underlying true shape has been proposed to simplify grasp generation \cite{box-based}. In addition, making use of local symmetry properties of objects has been shown to capture key shape features and generate heuristics based grasp candidates \cite{shape_approximation}, which however requires full observation of objects. Shape modelling, where objects are parameterized using smooth differentiable functions from point clouds via a spectral analysis \cite{grasp_moduli_spaces} has been employed to represent objects and grasps jointly in a common space allowing for transferring grasps on various objects. However this smooth parametrization can deteriorate with partial point cloud data, the shape space needs to accommodate missing data while avoiding unrealistic shape reconstructions. An alternative approach to explicit shape modeling is based on maximizing the contact surface between the object and the hand’s surface to find enveloping grasps, where precision grasping is not applicable that can be used for dexterous manipulation \cite{surface_match}. 

Touch sensing has also been used for understanding object shape, e.g. multi-finger tactile exploration using dynamic potential field \cite{bierbaum2009grasp}, which is used for extracting geometric features for grasping, tactile-servoing to explore surface features such as edges \cite{lepora2017exploratory}, 3D modelling to support vision where predicted shape from single camera image is refined with tactile sensing \cite{wang20183d}. Probabilistic approaches have also been studied for shape modelling and reconstruction. Probabilistic generative models of object geometry, trained with shape correspondences, have been used for inferring two-dimensional boundaries of partially occluded, deformable objects, and to recognize and retrieve complex objects from a pile for grasping tasks \cite{probabilistic_geometry}. However, these methods do not directly extend to processing views of three dimensional objects which contain multiple articulations and self-occlusions. Another promising shape modelling approach is GPIS, which has previously been used for: learning continuous sliding paths \cite{driess2017active}, single-finger tactile exploration to guide a sensor to high uncertainty regions \cite{yi2016active}, grasp planning in 2D using visual data only \cite{berkeley2DGPIS}, blind grasping by following surface contours for shape estimation \cite{stanimir_shape_grasp}, and grasping based on pre-trained systems using wrist poses\cite{graspGPIS}. In contrast, this paper addresses grasp planning with multi-finger hands in 3D given unknown objects, fusing visual and tactile data, and simultaneously optimizing grasp stability and enhancing perceived shape model without exhaustive exploration. We show how to iteratively improve grasp stability, by successive finger re-positioning, while also refining shape estimation.

Deep learning approaches are popular for predicting stable grasps given partially observed objects \cite{SurveyMultifingerDeepLearning, unknown_deep, lu2020multi}. They require training with grasp data extracted from a selection of training objects, which may sometimes fail to generalize well to new objects that have different properties from those trained on. A promising alternative to find stable grasp configurations for novel objects is through exploratory actions \cite{stanimir_shape_grasp, stanimir_grasp_explore}. Exploration-based approaches for grasping focus on finding grasp configurations without relying on expert knowledge or heuristics. \cite{stanimir_grasp_explore} follows an exhaustive exploration strategy using tactile data for shape estimation while applying different grasps. \cite{zito} chooses successive reach-to-grasp trajectories which maximise tactile information gain during iterative grasping attempts, in order to refine uncertain pose estimations for an object for which size and shape  are known a-priori. In \cite{sommer2016multi}, a control strategy for continuous exploration and grasping is proposed by exploiting the null-space of a multi-fingered hand. In \cite{graspBO}, a BO-based search is proposed, which however relies on complete object models and optimizing only wrist poses. 

Tactile sensing has been used for improving grasp planners e.g. using reinforcement learning to support open loop systems \cite{adaptivetactile}. In \cite{tactileregrasp}, a grasp adjustment approach was proposed, based on learning from tactile sensing alone, focusing on local re-grasping to improve stability, but without considering 3D geometry of the object. Differently from these works, we focus on improving the placement of the fingers of a multi-finger hand in terms of optimising grasp stability, while simultaneously exploring the shape of a partially unknown object, augmenting initial visual data with additional data from tactile glances.
%
\section{Bayesian Optimisation for Grasp Planning}
\label{sec:method}
This section presents the methodology for obtaining stable grasps and shape reconstruction based on the integration of visual perception with exploratory touching actions. The search for optimal grasps is performed by means of Bayesian Optimisation (BO), which is used to optimize functions that are either unknown or expensive to evaluate \cite{boTutorial}. For grasp planning, it is a convenient tool for handling complexity and uncertainty in various parameters. It searches for the extremities of an unknown objective function from which samples can be obtained.

BO is mainly comprised of two components used iteratively, (i) a statistical model to describe the observations of an objective function---the surrogate model---and (ii) an acquisition function that decides where to sample next  guiding the search to the optimum, while balancing the exploration vs exploitation trade-off. Gaussian Process Regression (GPR) is often used to create a statistical model (surrogate) which will be detailed in Section \ref{subsec:GPR}. In Section \ref{subsec:acquisition}, we present the acquisition function. We focus on maximizing grasp stability and the target function is the quality of a given grasp configuration. We use a 
probabilistic grasp quality metric, Probability of Force Closure (PFC), a robust metric that deals with the uncertainties inherent to the complex task of grasping unknown objects detailed in Section \ref{subsec:PFC}. Furthermore, explicit information about the perceived object shape to improve grasp search is added to the target function based on a GPIS model from visual and tactile observations, described in Section \ref{subsec:Unknown Shape Model}. The resulting target function combining both PFC and information from GPIS model is discussed in Section \ref{sec:surface_exploration}. 
In summary, let $f$ ($f:X \rightarrow \mathbb{R}, X\subset \mathbb{R}^d $, where $d$ is the dimensionality of the input space that is considered) be the real function to be optimised. Then, after placing a GP prior over $f$, the acquisition function is optimized based on the current surrogate model, which generates a novel query point $\bf x_i$ (that defines a grasp configuration as the locations of fingertips and the wrist). The grasp defined by $\bf x_i$ is executed and the corresponding grasp performance outcome $y_i$ is observed by means of evaluating the target function. The pair $\bf x_i$ and $y_i$ is thereafter used to update the surrogate model (posterior distribution) of $f$. These steps of choosing a new query point, augmenting the data with new observations and, updating the surrogate model is repeated. In Section \ref{sec:surface_exploration}, we present a detailed overview of our pipeline. 
\subsection{Gaussian Process Regression} \label{subsec:GPR}
Gaussian process regression (GPR) is a powerful tool to deal with regression problems, as it allows prior knowledge to be leveraged in order to estimate a smooth distribution of functions over the data, meanwhile providing uncertainty estimates. Formally, a Gaussian process (GP) can be a collection of $N$ random variables which have a joint Gaussian distribution, and therefore, can be completely specified by its mean and co-variance functions. Given a set of input points $\bm{X}=\left\{ \bm{x}_{1},\bm{x}_{2},\ldots,\bm{x}_{N}\right\} $ and observations $ \bm{Y}=\left\{ y_{1},y_{2},\ldots,y_{N}\right\} $, such that $ y_{i}=f(\bm{x}_{i})+\epsilon$, where $\epsilon\sim\mathcal{N}(0,\sigma_{\epsilon}^{2}$) denotes Gaussian noise with zero mean and $\sigma_{\epsilon}^{2}$ variance, the GP can be written as $f(\bm{x})\sim\mathcal{GP}\left(\ m(\bm{x}),k(\bm{x},\bm{x}')\ \right)$, where, $m(\bm{x})$ is the mean function and $k(\bm{x},\bm{x}')$ is the kernel or covariance function \cite{GP_Rasmussen}. Thus, given the kernel, the data, the predictive mean $\bar{f}(\bm{x}^{*})$ and variance $\mathbb{V}(\bm{x}^{*})$ at a query point $\bm{x}^{*}$ are%
\begin{align}
\bar{f}(\bm{x}^{*}) & = \mathbb{E}\left[f(\bm{x}^{*})\left|\bm{X},\bm{Y},\bm{x}^{*}\right.\right]=k(\bm{X},\bm{x}^{*})^{T}\bm{\Sigma}\bm{Y}   \nonumber 
\\
\mathbb{V}(\bm{x}^{*}) & = k(\bm{x}^{*},\bm{x}^{*})-k(\bm{x}^{*},\bm{x})^{T}\bm{\Sigma}k(\bm{x}^{*},\bm{x})^{T}\label{eq:GP_cov_and_mean}
\end{align}%
with $\bm{\Sigma}=\left(k(\bm{X},\bm{X})+\sigma_{\epsilon}^{2}\bm{I}\right)^{-1}$. In this work, GPR is employed to model both the BO surrogate function and the implicit surface representation of objects. For the surrogate model in BO, we use the squared exponential kernel which yielded a good performance in our experiments, $k_{\text{SE}}(\bm{x},\bm{x}')=\sigma_{\text{SE}}^2\exp\left( {-\tfrac{r^2}{2l^2}} \right)$ with $r=\left\Vert \bm{x}-\bm{x}'\right\Vert$, $\sigma_{\text{SE}}=0.001$ and $l=1$.
\begin{figure*}
\centering
\includegraphics[width=\textwidth]{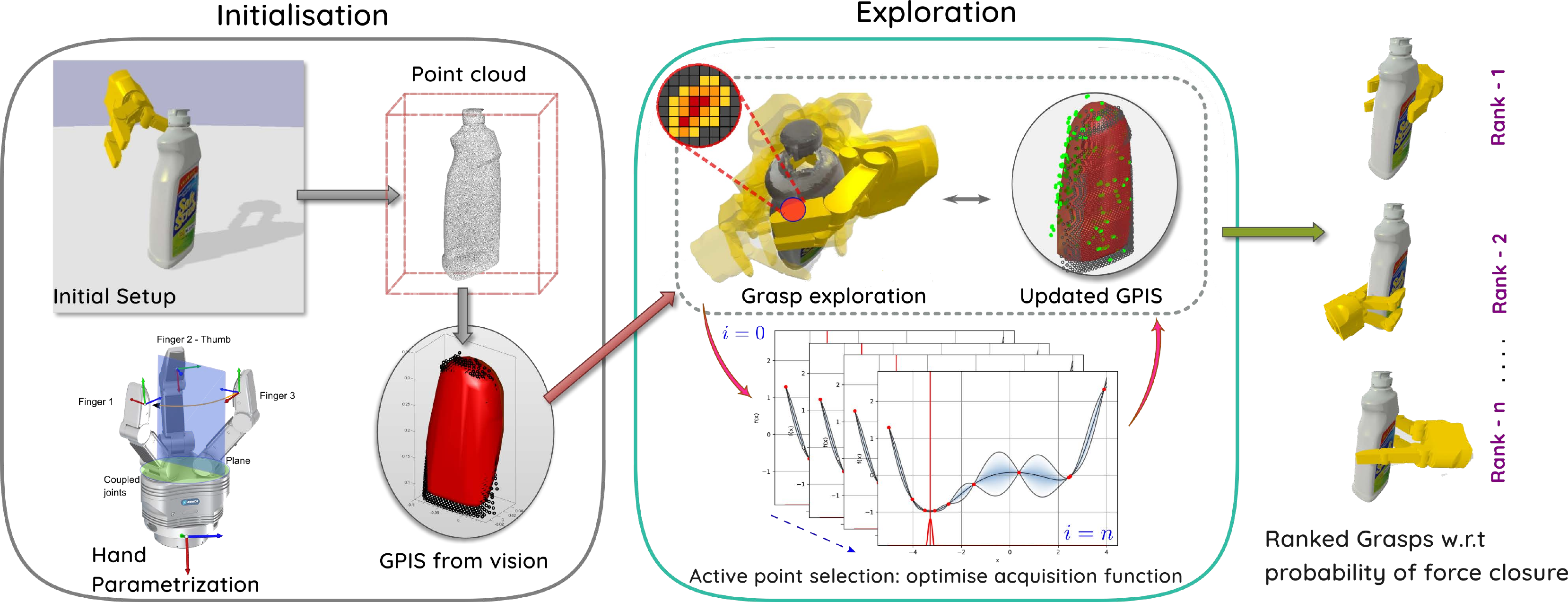}
\caption{BO-based exploration pipeline for shape reconstruction and grasp search.}
\label{fig:BOalg}
\end{figure*}
\subsection{Acquisition Function} \label{subsec:acquisition}
The acquisition function is used to select the next query point. Let $f(\bm{x})$ be the real objective function we wish to evaluate and,  $\bm{X}\in \mathbb{R}^{M\times N}$ and $\bm{Y}\in \mathbb{R}^N$ are respectively the input/output pair of observations acquired, where $M$ is the input dimensionality and $N$ is the number of observations. Given the surrogate model derived from $\left[\boldsymbol X, \bm{Y} \right ]$ and \eqref{eq:GP_cov_and_mean} with the squared exponential kernel, we select the next point by optimising an acquisition function, the Expected Improvement (EI), defined as
\begin{equation}
   \text{EI}(\bm{x})=\left(\bar{f}(\bm{x})-{y^{best}}\right)\varphi\left( \alpha \right)+  \mathbb{V}(\bm{x}) \Phi\left( \alpha \right)
\label{eq:expected improvement}
\end{equation}%
where, $ \alpha = \tfrac{\bar{f}(\bm{x})-{y^{best}}}{\mathbb{V}(\bm{x})} $, $y^{best}$ is the best sample so far, $ \Phi \left(\cdot\right)$ is the standard probability density function and $\varphi\left(\cdot\right)$ is the standard cumulative distribution function. Intuitively, EI can be maximised by optimizing two main elements of \eqref{eq:expected improvement}, \textit{i.e.}, $\left(\bar{f}(\bm{x})-y^{best}\right)$ and $\mathbb{V}(\bm{x})$. Maximising $(\bar{f}(\bm{x})-y^{best})$ means the next observation point will be where we expect the value will improve the most, and maximising $\mathbb{V}(\bm{x})$ means that the next observation should be made around the points we have less information about, which should refine our model and allow for a more informed decision about where the optimum point is. These two elements balance the main trade-off in the BO algorithms, exploration versus exploitation, with the EI function allowing us to explore the regions we know less about while at the same time looking for the highest difference between the current maximum and the rest of the function.
\subsection{Shape representation based on GPIS} \label{subsec:Unknown Shape Model}
While observing the data pair $\left[\boldsymbol X,\bm Y \right ]$, we also aim to reconstruct the surface of the object we are exploring so we can further use this information to guide our algorithm. We construct an implicit surface (IS) representation based on GPR \cite{microsoftPaper} as depicted in \eqref{eq:GP_cov_and_mean} to create a model of an unknown surface, while maintaining uncertainty about the reconstructed surface. The IS is formally defined as $f_{\text{IS}}(\bm{x}):\mathbb{R}^{3}\rightarrow\mathbb{R}$,
in which $f_{\text{IS}}(\bm{x})$ is the piece-wise function where $\bm{x}\in\mathbb{R}^{3}$ is the observed point:%
\begin{equation}
    f_{\text{IS}}(\bm{x})=
    \begin{cases}
        -1,\qquad &\text{if \ensuremath{\bm{x}} is below the surface}\\
        0,\qquad &\text{if \ensuremath{\bm{x}} is on the surface}\\
        1,\qquad &\text{if \ensuremath{\bm{x}} is above the surface.}
    \end{cases}
\label{eq:implicit surface}
\end{equation}
The IS is then modelled via GPR using vision and tactile points acquired from the camera and tactile sensors during object exploration.

To construct GPIS model of an object, we use the \textit{thin-plate} kernel given by $k_{\text{TP}}(\bm{x},\bm{x}')=2r^{3}-3Rr^{2}+R^{3}$, where $r=\left\Vert \bm{x}-\bm{x}'\right\Vert$ and $R$ is the maximum possible value of $r$ \cite{microsoftPaper}. Similarly to \cite{bjorkman} this kernel leads to better reconstruction performance, compared to other kernels, such as the squared exponential kernel. Finally, in order to guarantee the GPIS is both closed and bounded, we place additional points in the boundaries of the scene and in the data's centroid, according to \eqref{eq:implicit surface}.
\subsection{Probabilistic grasp quality} \label{subsec:PFC}
Besides guiding the exploration over unknown surfaces, we are also interested in ensuring high-quality grasps. To this extent, we follow a probabilistic approach for grasp quality based on force closure, which we aim to maximise during the exploration. Various grasp quality metrics have been proposed in the literature, most focusing on analytic approaches \cite{borst, forceClosure, forceClosure2}. A common metric, based on force closure is presented in \cite{ferrari1992planning}, which requires to compute the space spanned by the friction cones generated by all contacts, the Grasp Wrench Space (GWS). By approximating the GWS by its convex hull, the metric computes the quality of a grasp as the smallest wrench $\varepsilon_{\text{GWS}}$ that can break the grasp. However, this approach requires precise parameters, which are usually uncertain in realistic settings \cite{berkeley2DGPIS, chen2016UncertaintyGrasp, laskey2015uncertainty, allen}.

In this work, we consider four main sources of uncertainty: 1) contact normals ($\bm{n}$) are estimated from the evolving surface model. Furthermore due to e.g. mechanical and feedback errors, the robot will not always be able to approach the object with the desired trajectory to close the fingers along the estimated surface normals. 2) Achieved contact positions ($\bm{c}$) may be different than the planned ones, due to e.g., unmodeled dynamics, imprecise kinematics of the robotic arm, errors in estimating the surface of the object. 3) The friction coefficient $\mu$ which affects the stability of a grasp is often unknown, depending on the material of the object and the finger surfaces. 4) The centre of mass ($\bm{p}_\text{com}$) is calculated from the GPIS based object model, that has an associated variance for the estimations. We model uncertainties regarding these parameters based on the assumption that they are sampled from Gaussian distributions: \textit{i.e.}, 
$\bm{n}\sim\mathcal{N}\left(\hat{\bm{n}},\sigma_{{\bm{n}}}^{2}\right)$, $\bm{c}\sim\mathcal{N}\left(\hat{\bm{c}},\sigma_{{\bm{c}}}^{2}\right)$, $\mu\sim\mathcal{N}\left(\hat{\mu},\sigma_{{\mu}}^{2}\right)$,  ${\bm{p}_\text{com}}\sim\mathcal{N}\left(\hat{\bm{p}}_\text{com}, \sigma_{\text{com}}^{2}\right)$, where $\hat{\bm{n}}, \hat{\bm c}, \hat{\bm p}_{\text{com}} \text{ and } \hat{\mu} $ are the means and $\sigma_{\bm n}, \sigma_{\bm c}, \sigma_{\text{com}} \text{ and } \sigma_{\mu}$ are the standard deviations of the normal distributions, respectively. Considering all these sources of uncertainty, the probabilistic force closure ($P_{FC}$) is defined as $ P_{\text{FC}}=P(\varepsilon_{\text{GWS}}>\delta)$ where $\delta > 0$ is a threshold that is set empirically. 
\subsection{Surface exploration and exploitation}
\label{sec:surface_exploration}
In this section, we present our approach of tactile exploration for reconstructing object surface models and finding stable grasp configurations using a three fingered dexterous hand. We complement visual data with tactile observations to enhance perceived object shape, which leads to better grasp planning in terms of finding higher quality grasps.

Fig. \ref{fig:BOalg} depicts the overall pipeline of our work. It consists of three main components: initialisation, BO-based tactile exploration and post-exploration evaluation. During the pre-exploration phase, the system is initialized with the hand placed in its initial home configuration, \textit{i.e.}, placed at a distance above the object with fingers open. Using an RGB-D camera mounted in the scene facing the object, the point cloud of the object is captured, which corresponds to a partial view of the object visible to the camera. To choose the query points, a bounding box is placed around the object point cloud to define the boundaries of the domain, $\mathcal{S}\subset\mathbb{R}^M$, where $M$ is the dimensionality of the input. The initial GPIS is constructed using the vision information. After the main BO algorithm starts the GPIS model is updated with new tactile observations and $P_{FC}$ is monitored to find good grasp configurations. 

After initialisation, the process starts with the acquisition function suggesting a new point to explore. Our work differs from previous work using BO for grasp planning \cite{graspBO, graspBO3D} in that we combine visual and tactile sensing to reconstruct object surface while searching for grasp configurations and considering uncertainties. We perform exploration using all degrees of freedom of the hand, not only the wrist position, which is done in the robot's 6D task-space. For three-finger Schunk SDH hand, the query point is the desired 3D global position for the thumb and the first finger. Due to the constraints imposed by the coupled SDH joint, the last finger's position will be in the 2D plane relative to its knuckle. In total, the number of dimensions is $M=12$, including the Euler angles for the wrist orientation (as the wrist position is constrained by chosen fingertip locations) and an offset distance along the approach vector for the wrist, which allows it to move closer to the object. The hand is positioned accordingly for each query point and model is updated with resulting observations after closing the fingers until contact.

While seeking good grasps with high $P_{FC}$ and improving perception of the object shape, we also aim to establish contacts on the object surface during exploration and penalize the cases where there is no contact. To realize this, the target function can be defined as $y({\bar{f}_{\text{IS}}({\bm{x}_{F_i}}), \text{P}_{FC}})$ in which $\bar{f}_{\text{IS}}({\bm{x}_{F_i}})$ is obtained from the GPIS model given the fingertip positions for each finger $F_i$ ($i=1\ldots N_F$; for Schunk hand $N_F = 3$):
\begin{equation}
    y\left ( \bar{f}_{\text{IS}}({\bm{x}_{F_i}} \right ), \text{P}_{FC}) = \lambda\text{P}_{FC} - {\sum}_{i=1}^{N_F} \left [\bar{f}_{\text{IS}}({\bm{x}_{F_i}})\right ]^2
    \label{eq:target_function_2}
\end{equation}
where, we include both the probability of force closure to maximize (scaled by a constant, $\lambda$, which is set empirically), and a penalty term to guarantee that observations will be close to GPIS surface model, via the predictive mean of the GPIS in all contact points ($\sum_{i=1}^{N_F} \left [\bar{f}_{\text{IS}}({\bm{x}_{F_i}})\right ]^2$). This term yields for points on the surface $\bar{f}_{\text{IS}}({\bm{x}_{F_i}}) \approx 0$, whereas contact points moving away from the surface will quickly grow, which are then penalized. As new samples are obtained, GPIS and its center of mass are updated, the surrogate model uses the new observations and this process can be repeated until the number of desired iterations or the number of grasps is reached. Finally, during post-exploration evaluation, grasps with the highest $P_{FC}$ are evaluated by a number of tests, \textit{i.e.}, through lifting and shaking tests as in \cite{BenchmarkERL}. 

We summarise our method in Algorithm \ref{alg:Bayesain-Optimisation-Exploration}. The algorithm is initialised with the following: the GPIS model generated from the point cloud from the camera and its initial centre of mass ($\bm{p}^0_\text{com}$, as an additional data point as mentioned in Section \ref{subsec:Unknown Shape Model}), mean and variance for friction coefficient ($\hat{\mu}$ and $\sigma_{{\bm{\mu}}}^{2}$), the variance for normals ($\sigma_{{\bm{n}}}^{2}$), the variance for the contacts (${\sigma_{\bm{c}}}^{2}$) and the priors, which are obtained from Gaussian samples before the exploration algorithm starts. Further details on the optimisation variables are explained in Section \ref{sec:experiments}. Additional parameters of the algorithm are the number of samples $N_{S}$ for each variable to address their associated uncertainty, and, the number of iterations $N_{\text{STOP}}$ to stop the algorithm. From input data, BO algorithm suggests a new query point $\bm{x}\in\mathcal{S}$ , we define $\bm x$ as the vector that includes all $M$ inputs. Based on $\bm{x}$ that includes the 3D position of each fingertip ($\bm x_{F_i}$) and wrist ($\bm{w}$), we use an inverse kinematics based controller to send the fingers and the wrist to their corresponding configurations, if reachable. The resulting joint configuration that the fingers reach is depicted by $\bm{\theta}$. From sensory feedback, we acquire mean observations regarding target contacts  $\hat{\bm{c}}_{i}$ and normals $\hat{\bm{n}}_{i}$ for each finger (indexed by $ i\in [1 \ldots N_F]$). If all fingers reach these contact areas, we obtain $N_{S}$ samples for ${\bm{n}}$, ${\bm{c}}$, ${\mu}$ and $\bm{p}_\text{com}$  from $\bm{n}\sim\mathcal{N}\left(\hat{\bm{n}},\sigma_{{\bm{n}}}^{2}\right)$, $\bm{c}\sim\mathcal{N}\left(\hat{\bm{c}},\sigma_{{\bm{c}}}^{2}\right)$, $\mu\sim\mathcal{N}\left(\hat{\mu},\sigma_{{\mu}}^{2}\right)$, $\bm{p}_\text{com}\sim\mathcal{N}\left(\hat{\bm{p}}_\text{com},\sigma_{\text{com}}^{2}\right)$. From these samples, we build the GWS and convex hull in order to calculate the probability of force closure ($P_{FC}$). If a finger does not detect any contact, we store the fingertip position, which is obtained based on the Forward Kinematics Model (FKM) at the reached joint configuration together with the wrist orientation. This configuration is recorded as an unstable grasp, that is, we set $P_{FC}=0$.
Finally, the GPIS and the surrogate model are updated with new observations which are passed to the next iterations, together with updated $\bm{p}^0_\text{com}$.  When a new iteration starts another query point is suggested. This process is repeated for a fixed number of iterations, $N_{\text{STOP}}$. The algorithm then returns a list of the $j$ wrist poses ($\bm{w}$) and hand joint configurations $\bm{\theta}$, namely the grasps with the probabilities of force closure that are larger than a threshold, \textit{i.e.}, 0.5.

\begin{algorithm}[t]
	\SetAlgoLined
    \SetKwInOut{Input}{input}\SetKwInOut{Output}{output}
	\Input{\small{GPIS, priors}, $\hat {\mu}$, $\sigma^2_{{\mu}}$, $\sigma^2_{{\bm{n}}}$,  $\sigma^2_{{\bm{c}}}$, $\bm{p}^0_\text{com}$, $N_{S}$, $N_{\text{STOP}}$}
    \Output{ \{$ \bm{w}^{1...j}, \bm{\theta}^{1...j} $\} }
    \While{TRUE}
    {
        $\bm{x} \leftarrow $ EI in \eqref{eq:expected improvement}; $ \bm{w}, \bm{\theta}  \leftarrow \bm{x} $\\
        $ \hat{\bm{c}}_{i}, \hat{\bm{n}}_{i} \leftarrow $ sensory feedback, $\quad i\in [1 \ldots N_F]$ \\
        \eIf{$ \bm \hat{\bm{c}}_{i}\not=$NULL : $\forall  \hat{\bm{c}}_{i}$}
        {
            $\bm {x}_{F_i} \leftarrow \bm{\hat{c}}_{i} \quad i\in [1\ldots{N_F}] $\\
            $P_{\text{FC}} {\leftarrow} \{\bm{a}_1{\ldots}{\bm{a}}_{N_S}\} {\sim} \mathcal{N}{\left(\hat{\bm{a}},\sigma_{{\bm{a}}}^{2}\right)}, {\forall} \bm{a} {\in} \{ {\bm{n}}, {\bm{c}}, {\mu}, {\bm{p}}_\text{com}\}$\\
        }
        {
            \eIf{$\hat{\bm{c}}_{i}=$NULL}
            {
                $\bm{x}_{F_i} \leftarrow \text{\footnotesize{FKM}}(\bm{\theta}), i\in [1\ldots{N_F}]$\\ $P_{\text{FC}} \leftarrow 0$
            }
            {
                $\bm{x}_{F_i} \leftarrow \bm{\hat{c}}_{i} \quad i\in [1\ldots{N_F}]$ 
            }
        }
        $\bar{f}_{\text{IS}} ({\bm{x}_{F_i}})$  $\leftarrow $ \eqref{eq:GP_cov_and_mean}; $y (\bar{f}_{\text{IS}} ({\bm{x}_{F_i}}), P_{FC} ) \leftarrow $  \eqref{eq:target_function_2}\\
        Update the surrogate with ($\bm{x}_{F_i}, y$), \small{GPIS with} $\bm{x}_{F_i}$ \\
        Update $\bm{p}_\text{com}$ from GPIS\\
        \If{$N_{\text{STOP}}$ is reached}
        {
        \If{$P_{FC}>0.5$}
        {
            $ \bm{w}^j, \bm{\theta}^j  \leftarrow  \bm{x}_{F_i} $}
            \textbf{break}
        }
    }
	\caption{Exploration and grasp refinement.}
	\label{alg:Bayesain-Optimisation-Exploration}
\end{algorithm}
%
\section{Experimental Validations}
\label{sec:experiments}
In this section we present the experiments conducted in a simulation environment, Pybullet. Our robot setup consists of a three finger Shunk SDH2 hand and a 7 degrees of freedom arm, Kuka iiwa. Proof of concept of our approach has been demonstrated with a free-floating hand and also including the arm in explorations. Furthermore grasp evaluations are performed using the hand mounted on the robot arm. We validate the efficiency of our approach by comparing its performance to a baseline method whose details are provided below. For experiments, we have used seven different object models from two publicly available benchmark object sets. Five are from YCB object set \cite{ycb} and the remaining two are from CapriDB \cite{capridb} database. 
\subsection{Validating GPIS evolution}
\label{subsec:validation}
Initial exploration tests are conducted with a free-floating hand to explore the space without any reachability restrictions as well as to demonstrate our approach. Nevertheless, this does not limit our approach with a real setup. The exploration process is initialised with the GPIS constructed from a partial point cloud. Once the BO starts, the points in the task-space are selected to which we command the fingers. If the target position is not reachable, the closest possible location is selected instead. The finger movement is stopped once a contact is established. At this point we compute the average normal at the centre of the contacts and calculate the probability of force closure. Here to compute \eqref{eq:target_function_2} we use $\sigma_{{\boldsymbol{n}}}^{2}=\pi/8$, $\sigma_{{\boldsymbol{c}}}^{2}=0.0025$, $\sigma_{{\mu}}^{2}=0.1250 $, $\sigma_{{\text{com}}}$ which is obtained from the object model, $N_S=10$. 

\begin{figure*}
\centering
    \includegraphics[width=\textwidth]{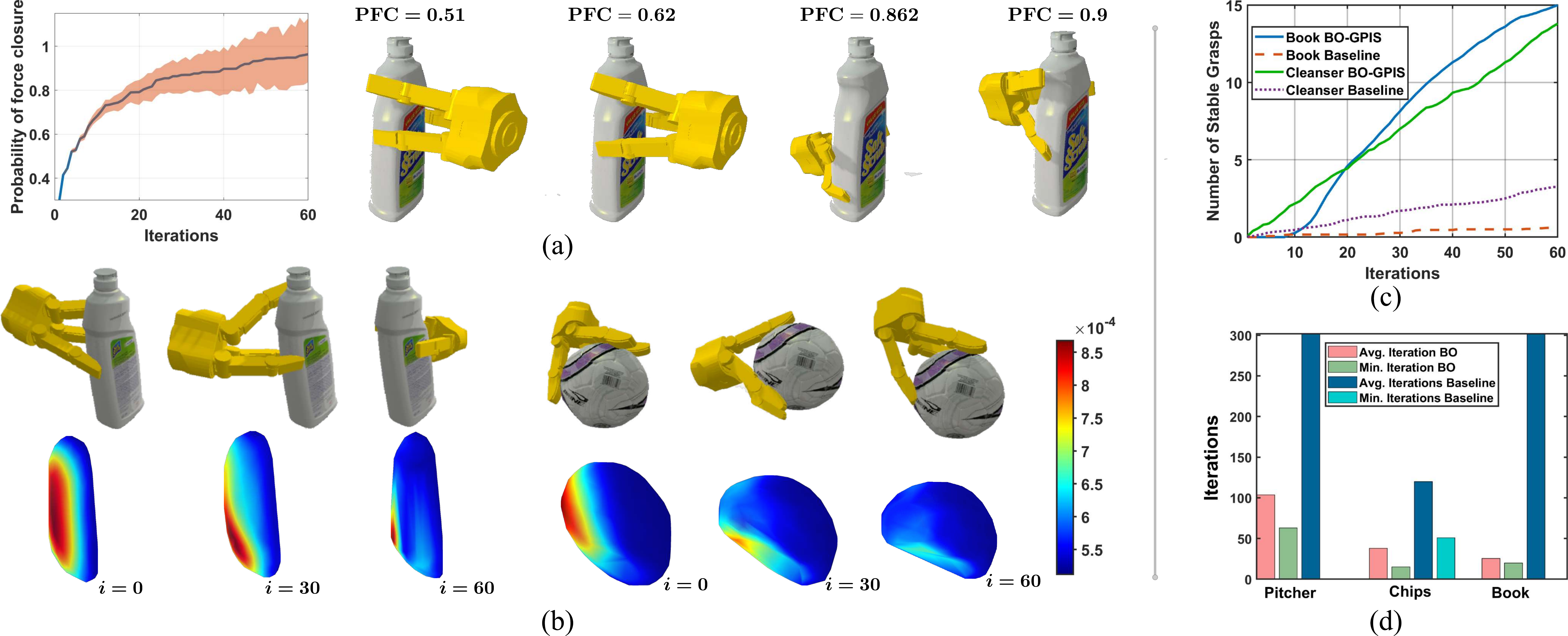}
    \caption{Experimental results. (a) evolution of best $P_{FC}$ over 30 experiments and a few examples taken from one of those exploration experiments; (b) model evolution for both "cleanser" and "ball" objects at iterations $i =0, 30, \text{ and } 60$ and corresponding grasps. Red regions in surface maps correspond to high uncertainty and blue to low; (c)  evaluation of the number of sable grasps compared to the baseline; and (d) evaluation of the number of iterations required to obtain $10$ grasps. More detailed results can be found in the attached supplementary video.}
    \label{fig:GPISBO_REsults}
\end{figure*}

As illustrated in Fig.~\ref{fig:GPISBO_REsults}a, our approach progressively improves the quality of the initial grasp configuration as we iterate through the optimization loop, where $P_{FC}$ increases during exploration steps and grasp configurations become more stable. The perceived object shape evolves to resemble the true object as tactile glances are added as in Fig.~\ref{fig:GPISBO_REsults}b where the uncertainty in GPIS decreases in the explored areas.

\begin{figure}
\centering
    \includegraphics[width=\columnwidth]{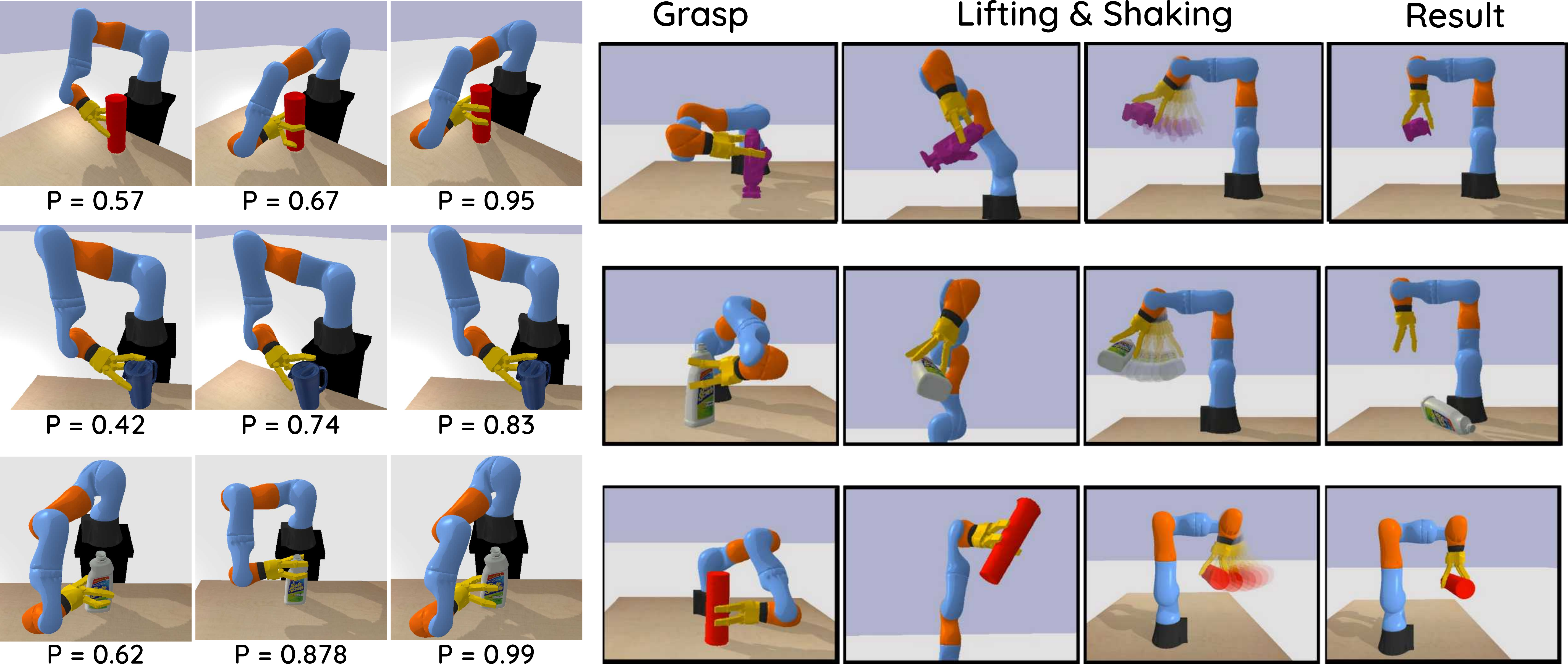}
    \caption{Exploration and test examples (lifting and shaking) using an arm.}
    \label{fig:GPISBO_FullArmExp}
\end{figure}

In Table \ref{tab:merged_table_noarm},  we present the results obtained for a list of objects. For these experiments, the optimization loop is initialized with a prior obtained from Gaussian samples corresponding to finger positions from pre-exploration grasps. The number of these grasps depends on the object complexity and we build the priors based on at least 3 grasps with $P_{FC}$ values $>0.4$. The exploration experiments presented in Table \ref{tab:merged_table_noarm} are run with $N_{\text{STOP}}=60$  and are repeated 30 times for each object. We report average of best $P_{FC}$ values, average number of stable grasps whose $P_{FC}$ $>0.5$ and their average success rates from every trial. Success rates are computed based on lifting and shaking experiments where the object is first lifted $20~\textrm{cm}$ above the table and then a series of shaking movements (following a sinusoidal trajectory with frequency $0.4~\textrm{Hz}$ at $0.59~\textrm{rad/s}$) are applied \cite{BenchmarkERL}. Grasp configurations leading to successful lifting and shaking tests are labeled successful. It can be seen from the table that the approach leads to stable grasp configurations with high $P_{FC}$ and high lifting rates. However, a fall in the success rates after shaking test is observed for some objects. We believe this is due to uncertainties in the physics models of the objects and fingers, e.g. imprecise object collision models, physics engine uncertainties etc. We highlight that we imposed harsh conditions to the shaking test, e.g., high accelerations. We also validate our approach by using a KUKA arm for exploration experiments as seen in Fig. \ref{fig:GPISBO_FullArmExp}. This results in grasp configurations that are reachable for the arm and, exploration that is still able to increase the probability of force closure. Fig. \ref{fig:GPISBO_FullArmExp} also shows example lifting and shaking  experiment results.
\midsepremove
\begin{table*}
\smaller
    \centering
        \caption{Experimental evaluation with multiple objects using proposed approach.}
        \label{tab:merged_table_noarm}
\begin{threeparttable}
    \renewcommand{\tabularxcolumn}[1]{m{#1}}
\begin{tabularx}{\textwidth}{>{\hsize=0.1\hsize\raggedright\arraybackslash}X|
                              >{\hsize=0.1\hsize\centering\arraybackslash}X|
                              >{\hsize=0.1\hsize\centering\arraybackslash}X|
                             >{\hsize=0.1\hsize\centering\arraybackslash}X|
                             >{\hsize=0.15\hsize\centering\arraybackslash}X|
                             >{\hsize=0.15\hsize\centering\arraybackslash}X|
                              >{\hsize=0.05\hsize\centering\arraybackslash}X|
                              >{\hsize=0.05\hsize\centering\arraybackslash}X|
                             >{\hsize=0.05\hsize\centering\arraybackslash}X|
                             >{\hsize=0.05\hsize\centering\arraybackslash}X}
\toprule
 \multirow{2}{*}{Object\tnote{1}} & \multicolumn{2}{c|}{Best Prob. (Std.)} & GPIS Evolution\tnote{2}\footnotesize{[\%]} & \multicolumn{2}{c|}{Total Grasps (Avg., Std.)} &  \multicolumn{2}{c|}{Lifting  \%} & \multicolumn{2}{c}{Shaking \%}   \\
   \cmidrule{2-10}
       & BO & Baseline & BO  & BO & Baseline & BO & Baseline & BO & Baseline \\
     \midrule
     Pitcher & {$1.0 (0.0)$} & {$0.82 (0.14)$} & {$18.56, 1.59$} & {$469{(15.6, 8.23)}$} & $117{(3.90, 2.06)}$ & {$80.19$} & {$86.06$} & {$ 51.78$} & {$ 44.70$} \\
     Chips & {$1.0 (0.0)$} & $0.92 (0.08)$ & {$7.78, 3.9$} &{$792{(26.4, 10.6)}$} & $242{(8.07, 2.53)}$ & {$86.99$} & {$84.77$} & {$78.82$} & {$60.54$}\\
     Cleanser & {$0.99 (0.04)$}  & $0.81 (0.19)$ & {$20.40, 13.9$} &{$501{(16.7, 7.4)}$} & $101{(3.4, 1.90)}$ & {$ 86.48$} & {$84.25$} & {$ 71.72$} & {$ 68.62$} \\
      Ball & {$0.93 (0.14)$} & $0.71 (0.19)$ & {$2.02, 10.02 $} &{$148{(4.9, 2.6)}$} & $62{(2.07, 1.36)}$ &{$72.46$} & {$68.07$} &{$60.4$}& {$ 46.73$} \\
     Woodblocks & {$0.95(0.14)$} & $0.55(0.25)$ & {$15.02, 1.06$}&{$440{(14.7, 11.2)}$}  & $25{(0.83, 1.02)}$  &{$78.74$} & {$100$} &  {$32.23$} & {$ 44.87$} \\
      Book & {$0.98 (0.07)$} & $0.48 (0.38)$  & {$27.3, 15.53$}  &{$266{(8.87, 6.0)}$} & $19{(0.63, 0.72)}$ & {$92.76$} & {$38.89$} &{$ 80.36$}& {$ 22.22$} \\
      Toy Robot & {$1.0 (0.01)$} & $0.83 (0.18)$ & {$24.69, 1.07 $} & {$375{(12.5, 5.4)}$} & $87{(2.90, 1.79)}$ & {$ 86.54$} & {$61.63$} & {$ 77.14$} & {$ 59.30 $}\\ 
   \bottomrule
\end{tabularx}
\begin{tablenotes}
\item[1] First five are from YCB object set \cite{ycb} and the last two are from CapriDB\cite{capridb} database.
\item[2] Rate of improvement of the Hausdorff distance from ground truth and rate of change of variance.
\end{tablenotes}
\end{threeparttable}
\end{table*}
\subsection{Baseline comparisons}
We compare our approach with a point cloud-based heuristic exploration baseline in which we run exploration trials for 60 iterations. Here, instead of BO guiding the fingers, we draw samples from a Gaussian in the same domain as the BO. To facilitate this heuristic, we restrict the thumb position to the neighbourhood of randomly sampled location from the point cloud with $\sigma_{th}=0.05$. Table~\ref{tab:merged_table_noarm} presents the results for baseline experiments applied to all test objects. It can be noticed that the stable grasps and $P_{FC}$ values are reduced in comparison to our approach. Our method outputs a larger number of grasps satisfying both lifting and shaking tests. Notice that even the percentage of success is greater with the only exceptions being the woodblocks (where the baseline finds few stable grasps) and the lifting test for pitcher which still corresponds to larger number of successful cases. These results clearly demonstrate the efficiency of our proposed method.
\subsection{Discussion}
It can be seen in Table \ref{tab:merged_table_noarm} that our approach increases both the best probability and the average number of stable grasps. The advantages of our method are best illustrated with experiments using the book model. It particularly highlights one of our method's strengths: the ability to search around optimal values. In the baseline, positions are sampled randomly around the object, whereas good grasps are only possible on the thinner parts of the book. Hence, it rarely finds successful results. Our method, on the other hand, uses the previous explorations to guide the hand and grasp where it is more likely to succeed. Most of the exploration happens in that vicinity, facilitating local improvements. %
Fig.~\ref{fig:GPISBO_REsults}c  shows the evolution of average number of stable grasps for two objects: book and cleanser. As there are less good grasp regions for the book, few grasps are found for the baseline whereas our approach quickly finds optimal regions. The cleanser has more good grasp regions which are feasible, thus it is easier for the baseline to find stable grasps. Nonetheless, our method converges to the optimal regions. It also finds more grasps quickly.

Fig.~\ref{fig:GPISBO_REsults}d shows how our approach can find grasps rapidly compared to the baseline. It shows the number of iterations needed for each method to find 10 grasps with $\textrm{$P_{FC}$}>0.8$. These experiments are conducted 5 times for each object up to a maximum of 300 iterations. If less than 10 grasps are found, we consider it as no convergence possible. For three objects shown in Fig.~\ref{fig:GPISBO_REsults}d, the baseline was only successful with the chips can, with an average of 120 iterations (minimum 51) to finish the experiment. In comparison, our method only required an average of 37.8 iterations (minimum 15) for the chips can. For pitcher, using our approach, the experiments finished at an average of 103.6 iterations (minimum 63), and for book the average was 25.4 iterations (minimum 20). Table~\ref{tab:merged_table_noarm} also shows the GPIS mean and variance evolution per object with the tactile exploration, as the difference between the distance to the ground truth in the beginning and at the end of the exploration, and the drop in the variance of GPIS. We observe the average distances to the ground truth and the surface uncertainty decrease during exploration.
%
\section{Conclusion}
\label{sec:conclusion}
This paper has presented an exploration-based grasping method for unknown objects using a multi-finger hand based on visual and tactile data. Our approach is based on BO which provides a principled way to search for parameters that maximize grasp stability while improving perception of object shape. Exploration starts by initially only using visual data, then it relies on tactile sensing to search for better configurations. We make use of a probabilistic force closure-based evaluation of grasp configurations to account for uncertainties in relevant parameters such as shape, friction, contact points and normals. Results show that the proposed approach reaches stable grasp configurations in less iterations than the baseline. We will validate our approach on an identical real setup directly, as the method is well-suited to address perception uncertainties that would arise in a real setup. We plan to extend the work in different directions: by improving the shape representation (e.g. using geometric priors), adding task constraints to grasp planning, extending the approach for grasp transfer and to bi-manual grasping, and by including simultaneous pose tracking to deal with non-stationary cases.
\bibliographystyle{IEEEtran}
\bibliography{references}
\end{document}